\documentclass[a4paper]{article}

\usepackage{spconf,amsmath,graphicx}

\title{Speech Recognition with Augmented Synthesized Speech}
\name{Andrew Rosenberg, Yu Zhang, Bhuvana Ramabhadran, Ye Jia, Pedro Moreno, Yonghui Wu, Zelin Wu}
\address{
  Google\\
{\small {\{rosenberg, ngyuzh, bhuv, jiaye, pedro, yonghui, zelinwu\}@google.com}}}

\begin{document}

\maketitle
\begin{abstract}
Recent success of the Tacotron speech synthesis architecture and its variants in producing natural sounding multi-speaker synthesized speech has raised the exciting possibility of replacing expensive, manually transcribed, domain-specific, human speech that is used to train speech recognizers. The multi-speaker speech synthesis architecture can learn latent embedding spaces of prosody, speaker and style variations derived from input acoustic representations thereby allowing for manipulation of the synthesized speech. In this paper, we evaluate the feasibility of enhancing speech recognition performance using speech synthesis using two corpora from different domains. We explore algorithms to provide the necessary acoustic and lexical diversity needed for robust speech recognition. Finally, we demonstrate the feasibility of this approach as a data augmentation strategy for domain-transfer. 
We find that improvements to speech recognition performance is achievable by augmenting training data with synthesized material.  However, there remains a substantial gap in performance between recognizers trained on human speech those trained on synthesized speech.


\end{abstract}

\section{Introduction}

Speech synthesis (text-to-speech or TTS) performance has improved in recent years.  These improvements have been so dramatic that, in some cases, synthesized speech is indistinguishable from human speech \cite{shen2018natural}.
One reliable way, perhaps even the most reliable way, to improve automatic speech recognition (ASR) performance is to add more transcribed training data.
If speech synthesis is equivalent to human speech, adding more transcribed training data generated by speech synthesis should improve speech recognition performance.
In this paper, we test this hypothesis. We explore this using data augmentation, where the human speech training data is combined with various amounts and types of synthesized material.  

Our experiments use two corpora.  {\sc LibriSpeech} is a 960 hour corpus of books read by non-professional speakers \cite{panayotov2015librispeech} divided into a 460 hour ``clean'' portion and a 500 hour ``other'' partition.  These partitions are created by speaker splits with those speakers whose speech is easy to recognize being put in the ``clean'' partition, and more difficult speakers in ``other''.  We also use {\sc Isolated-Sentences}, an internal corpus of shorter utterances read in diverse recording conditions.  {\sc Isolated-Sentences} contains 76 hours of material across 201k utterances collected from 1,988 speakers.  The number of speakers in this corpus is similar to {\sc LibriSpeech}, but the total material is significantly less.

The connection with speech synthesis and speech recognition has been explored previously (cf. Section \ref{sec:related-work}).  We draw particular comparison with \cite{li2018training}, prior work exploring speech synthesis based data augmentation on {\sc LibriSpeech}.

For speech synthesis, we use a Tacotron 2 \cite{shen2018natural} based multi-speaker speech synthesis model with a WaveRNN vocoder \cite{kalchbrenner2018efficient}.  Specifics of the model are described in Section \ref{sec:tts-model}.   We train slightly different version of the synthesizer whether training on {\sc LibriSpeech} or {\sc Isolated-Sentences}.
For speech recognition we use an end-to-end, encoder-decoder with attention recognizer (cf. Section \ref{sec:asr-model}).  We use the same recognition model for {\sc LibriSpeech} recognition and {\sc Isolated-Sentences} recognition.

We describe {\sc LibriSpeech} data augmentation experiments in Section \ref{sec:ls-data-augmentation}.
The synthesizer is capable of producing speech from multiple speakers based on a fixed sized speaker embedding.  We describe different approaches to controlling the speaker representation and report their impact on data augmentation performance in Section \ref{ssec:speaker-representation-augmentation}.  
We also investigate how ASR performance is impacted as we reduce the amount of human speech (cf. Section \ref{ssec:reduced-data}).  This seeks to address the question of whether training data can be replaced by synthesized material and with what impact on performance.  

Data augmentation is a well-established method to approximate noise and variability in training data to improve robustness and generalizability at evaluation or inference time \cite{cui2015data}.  Resynthesis of training utterances addresses speaker variability by generating multiple readings of the training utterances from different (synthesized) speakers.   However, this speaker variability is only part of the variability that speech recognition needs to be robust to; we expect a recognizer to recognize lexically diverse utterances.  To that end, we demonstrate the use of speech synthesis to include unseen utterances in the training data, resulting in a more robust recognizer (cf. Section \ref{sec:ls-lexical-diversity}).

Speech recognition (like any other machine learning application or statistical model) is most effective when the training data and testing data are drawn from the same population.  In the context of data augmentation, we ask if we can use a speech synthesis model trained on one domain, to synthesize material in a new domain and use this synthetic material to improve recognition performance.  If this is the case, a general purpose synthesizer can be used for data augmentation across a number of domains.  If it is not, speech-synthesis driven domain adaptation would require a domain specific synthesizer for each domain. This is limiting, as speech synthesis on ``found'' data is a challenging problem \cite{cooper2019text}.  To address this, we repeat the data augmentation and lexical diversity experiments performed on {\sc LibriSpeech} material on the {\sc Isolated-Sentences} domain, and investigate the relative value of a {\sc LibriSpeech}-trained synthesizer to one trained on {\sc Isolated-Sentences} (cf. Section \ref{sec:domain-adaptation}).

Overall, we find that expanded acoustic diversity and lexical diversity in training data as provided by synthesized speech can improve ASR performance.  However, there are limits to these improvements.  Despite closing much of the gap between synthesized and real speech with respect to synthesis quality, more work is needed for synthesized speech to deliver the value of real speech for ASR training.

To summarize, the contributions of this work are as follows:
\begin{enumerate}
    \item We demonstrate that data augmentation with TTS utterances yields improvements to ASR.
    \item We show, for the first time, the importance of effective and diverse speaker representations in TTS for ASR data augmentation.
    \item We show, for the first time, the impact of using lexically diverse TTS utterances for ASR data augmentation.
    \item We describe a novel TTS technique, hierarchical VAE, which significantly aids TTS training with a small amount of multi-speaker training data.
\end{enumerate}

\section{Related Work}
\label{sec:related-work}

The prior work most similar to the research described in this paper was written by Li et al.~\cite{li2018training}.  They also investigated the use of speech synthesis as a source for data augmentation.  We find similar gains via expanding acoustic diversity.  Li et al. accomplished this via a Global Style Token (GST) to expand prosodic variation as described in \cite{wang2018style}. They were also able to achieve additional performance improvements through hyper parameter tuning, specifically increased depth of the recognition network.  We confirm these findings through a distinct but related approach to expanding acoustic diversity of the training data, and expand on these via investigations of lexical diversity.  In terms of specific results on {\sc LibriSpeech}, our work yields better performance on the more challenging test-other partition, while Li et al. report better results on the test-clean partition.

A different approach that jointly trains ASR and TTS systems has been explored. The term, {\it speech chain}, was first introduced in 1963 (and reissued in 1993) by Denes et al.~\cite{denes1963speech,denes1993speech}, where  speech communication was described as a spoken message between the minds of the speakers and the listeners, based on speech production and speech perception processes. Using this as a basis, Tjandra et al., developed DeepChain~\cite{tjandra2017listening}, an approach to simultaneously train both recognition and synthesis systems. They proposed a sequence-to-sequence model in a closed-loop architecture that allows training with both, labeled and unlabeled data. The ASR component  transcribes the unlabeled speech features, that TTS synthesizes from, while the ASR component also attempts to learn from the labeled text using the synthesized speech. Multi-speaker, Tacotron TTS with speaker embeddings was used in this work. Using the BTEC corpus, the authors demonstrated an improvement in ASR performance when allowing TTS to learn from ASR and viceversa. This work was extended further in~\cite{tjandra2018machine} with the introduction of a speaker verification module inside the closed-loop architecture. This module generates a speaker embedding with one utterance of the target speaker and allows the DeepChain model to handle unseen speakers. Improvements to ASR performance were demonstrated  on the WSJ corpus when measuring character error rate (CER).

\section{Speech Synthesis Model}
\label{sec:tts-model}
\begin{figure}[tb]
 \centering
 \centerline{\includegraphics[width=\linewidth]{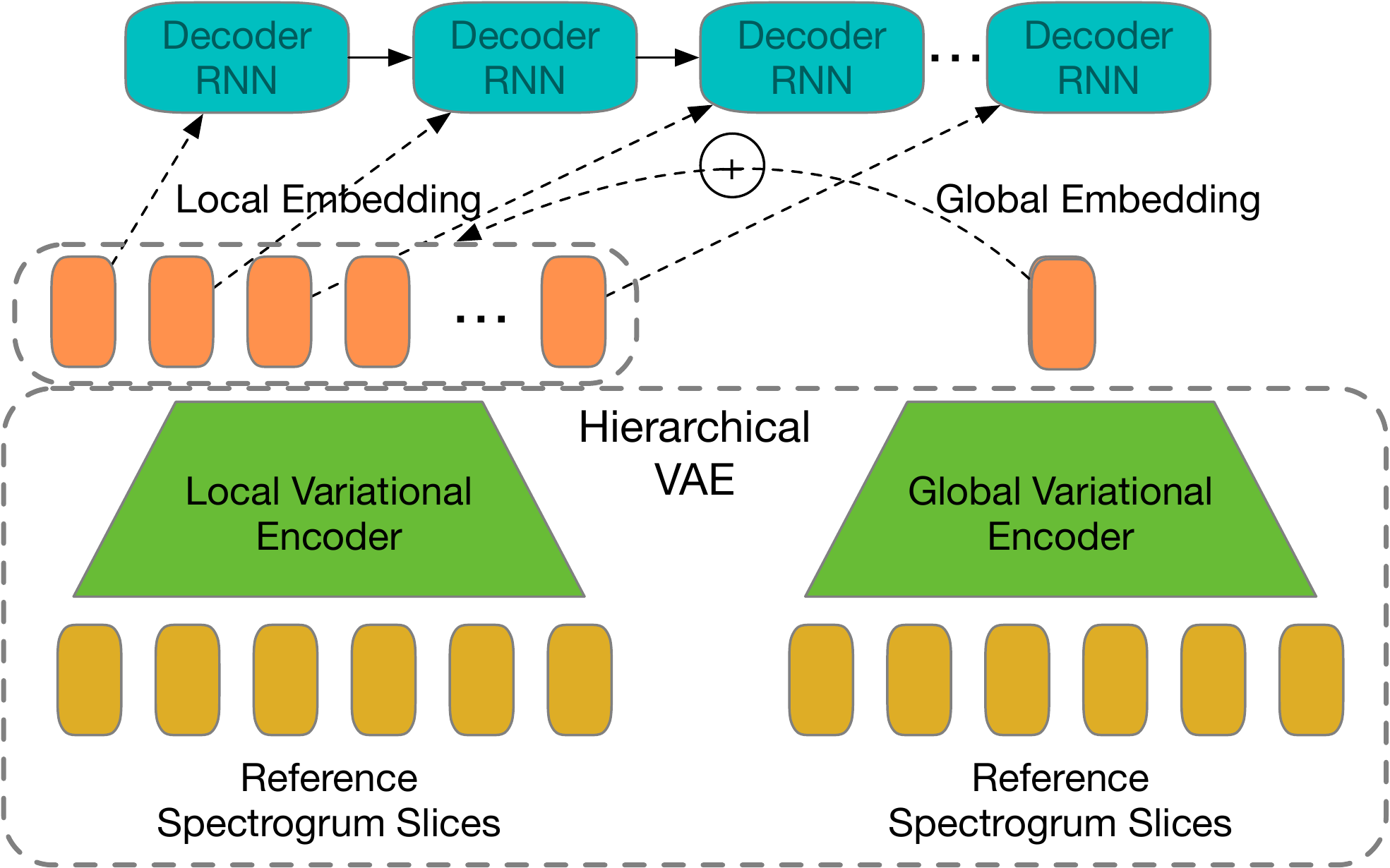}}
  \caption{
  Local VAE encodes local style information with fixed window size. Global VAE encodes global style information. The summed style vector is broadcast to each decoder RNN step.}
  \label{fig:hievae}
\end{figure}
We base our TTS model on Tacotron 2~\cite{shen2018natural}, which takes a text sequence as input, and outputs a sequence of mel spectrogram frames. The input text sequence embedding is encoded by three convolutional layers, which contain 512 filters with shape $5 \times 1$, followed by a bidirectional long short-term memory (LSTM) layer of 256 units for each direction. The resulting text encodings are accessed by the decoder through a location sensitive attention mechanism, which takes attention history into account when computing a normalized weight vector for aggregation.

The autoregressive decoder network takes as input the aggregated text encoding, and conditioned on 256-dimensional d-vector from a separately trained speaker encoder as in Tacotron 2D~\cite{jia2018transfer}. Similar to Tacotron 2, we separately train a WaveRNN~\cite{kalchbrenner2018efficient} to invert mel spectrograms to a time-domain waveform.  

Training TTS on non-studio data like Isolated-Sentences is challenging.  To improve TTS fidelity when training on this material we make two changes to the model.  First, we use a GMM based attention mechanism \cite{graves2013generating} rather than content-based attention.  Second, we augment the model with a variational auto encoder (VAE) as in \cite{hsu2018hierarchical}. In order to make the model robust to noise and highly biased data, we further modify the global VAE to a hierarchical version as Figure \ref{fig:hievae}. The new VAE includes a local encoder which encodes fixed two-second chunks with a one-second overlap and a global encoder which encodes the whole utterance. Both VAEs have convolutional layers followed by LSTM layers. The global style vector is summed with the local style vector and fed into each decoder RNN step. These two modifications add stability and fidelity to the TTS output. The use of a hierarchical structure in the VAE results in synthesized speech that is recognized by Google's production voice search recognizer with a WER that is 1\% absolute less than the non-hierarchical VAE. 

\section{Speech Recognition Model}
\label{sec:asr-model}

The ASR experiments in this paper use listen-attend-spell (LAS), an end-to-end encoder-decoder model with additive attention mechanism \cite{shen2019lingvo, chiu2018state}.   

Input speech is represented by 80-mel input features with deltas and double deltas.  These are  encoded by two convolutional layers, which contain 32 filters with shape $3 \times 1$ and a $2 \times 2$ stride, followed by four bidirectional LSTM layers of 1024 units for each direction.   The decoder is a two unidirectional LSTM layers with 1024 units.  Targets are drawn from a vocabulary of 16k graphemic word pieces \cite{chan2016latent}.  The word piece model inventory is learned from the {\sc LibriSpeech} training data.  These targets are used for both the {\sc LibriSpeech} and {\sc Isolated-Sentences} experiments. This model does not include a language model during training, and we do not perform any second-pass language model based rescoring.
Training is performed using Adam \cite{kingma2014adam} for 200k steps.  We use a warmup and exponential decay learning rate schedule with a maximum of 1e-3 and minimum of 1e-5 \cite{goyal2017accurate}.  We introduce variational weight noise with a standard deviation of 0.075 after 10k steps. 

This architecture, training strategy, and associated hyperparameters were tuned for {\sc LibriSpeech} ASR performance without any data augmentation.  While these settings may not always be optimal, we keep these fixed for all experiments. 

There are instances when end-to-end speech synthesis fails to faithfully synthesize the input utterance.  This typically comes from end-of-sequence issues, where the synthesizer either prematurely stops, or fails to end, and generates some ``babbling'' like noise.   This typically occurs on long utterances, and is a more significant issue when synthesizing {\sc LibriSpeech} material than {\sc Isolated-Sentences}.  Therefore, prior to training, we attempt to identify these poorly synthesized {\sc LibriSpeech} utterances. To do this, we decode them using a recognizer trained on {\sc LibriSpeech}.  Using the intended utterance, we eliminate synthesized utterances that are recognized with a WER of more than 20\%.  This eliminates between 10 and 20\% of utterances from training depending on the speaker representation used.  Because we have fewer end-pointing issues on short sequences, we do not do this filtering on {\sc Isolated-Sentences}.

For all experiments, we use the standard partitions of the {\sc LibriSpeech} material.  The training set comprises 460 hours of ``clean'' data, and 500 hours of ``other'' data.  Similarly dev and test sets are partitioned into ``clean'' and ``other'' subsets.  {\sc Isolated-Sentences} contains 76 hours of material, partitioned into train, dev and test sets with a 90/5/5 ratio.  For both corpora, the train, dev and test partitions remain constant across both TTS and ASR experiments.

\section{Acoustic Diversity}
\label{sec:ls-data-augmentation}

We describe data augmentation experiments on  {\sc LibriSpeech}.  In these,  training utterances are augmented with  duplicate synthesized copies of training text produced by TTS.

\subsection{Speaker Representation}
The multi-speaker TTS model described in Section \ref{sec:tts-model} uses d-vectors as speaker-conditioning information.  To control the speaker diversity in the synthesized data we use three different approaches to generate d-vectors for inference.\\
 {\bf Original} Use a d-vector derived from the training utterance itself.  In the case of perfect synthesis, this would result in a synthesized utterance that is identical to the source. \\
{\bf Sampled} Randomly select a d-vector from some other utterance that was used during training for inference.  This ensures that the speaker representation has been seen by the synthesizer, but the source utterance and synthesized utterance will differ in terms of the speaker characteristics.\\
{\bf Random} Generate a random 256 dimensional vector, and then project it to the unit-hypersphere via L2-normalization.  This is then used as the d-vector for the synthesized utterance.  If the d-vectors are evenly distributed 
this will result in an effective random sampling of speaker characteristics.

\subsection{ASR augmentation}
\label{ssec:speaker-representation}
\label{ssec:speaker-representation-augmentation}

Based on the three speaker representation approaches, we generate synthesized utterances for data augmentation. Each of these experiments include a complete synthesized copy of the human speech training data. ASR performance is reported in Table \ref{tbl:ls-asr-augmentation}.  The {\sc LibriSpeech} corpus contains two test (and development) partitions, one that is relatively clean (test-clean and dev-clean) and one that is not (test-other and dev-other).  
\begin{table}[htb]
    \centering
    \begin{tabular}{|c||c|c|}
    \hline
    Augmentation& test-clean & test-other \\
    \hline
         None&4.77&13.89  \\
         Original&4.84&14.75\\
         Random&4.92&14.91\\
         Sampled&4.58&13.78 \\
         \hline
    \end{tabular}
    \caption{{\sc LibriSpeech} Word Error Rate (WER) by augmentation with TTS utterances using different d-vector approaches.}
    \label{tbl:ls-asr-augmentation}
\end{table}
This reveals that acoustic diversity as controlled by the speaker representation used during synthesis has an impact on the value of synthesized speech for data augmentation.   Random d-vectors are less effective in generating effective speaker diversity.  This likely is because of a mismatch between the d-vectors used during inference and those seen during training.  The training d-vectors are not uniformly distributed on the unit hypersphere.  Synthesis quality suffers when the supplied d-vector is inconsistent with observed speaker conditioning information.  Using the original d-vectors does not provide reliable gains, and in fact, degrades performance somewhat.  Sampling from observed speaker representations provides  speaker diversity. This results in a relative gain of approximately 4\% to WER on both the test-clean and test-other sets.

While this improvement is modest, the fact that performance is improved at all speaks to the coherence between the multi-speaker TTS utterances and human speech, as well as the ability for d-vector speaker representation to reasonably capture some of the acoustic diversity expressed by different speakers.  For comparison, we repeated this experiment with augmentation material from high-quality single-speaker TTS using a traditional TTS frontend and WaveNet vocoder.  Even though single-speaker TTS quality is superior to the multi-speaker system used in Table \ref{tbl:ls-asr-augmentation}, the resultant ASR performance is dramatically worse, specifically, WERs of 13.55 and 24.84 on test-clean and test-other sets, respectively.  Thus, TTS quality is a necessary but not sufficient criteria for effective data augmentation. Multi-speaker TTS is necessary for effective data augmentation for ASR. 

\subsection{Reduced Source Material}
\label{ssec:reduced-data}
We next investigate the performance of data augmentation as a function of the amount of source training data.   All experiments use the {\bf sampled} approach to speaker conditioning. Table \ref{tbl:reduced-data} describes ASR performance using all 960 hours of synthesized material, with reduced amounts of source speech, as compared to performance using only the reduced source material.
\begin{table}[htb]
    \centering
    \begin{tabular}{|c||c|c||c|c|}
    \hline
    &\multicolumn{2}{c||}{data aug}&\multicolumn{2}{c|}{source only}\\
    hrs&clean&other&clean&other\\
    \hline
        0& 32.44 &66.10&-&-\\
         10-clean&16.91&43.10&NA&NA\\
         100-clean&9.25&30.58&12.46&34.00\\
         460-clean&6.28&22.52&6.30&22.41\\
         960-all&4.58&13.78&4.77&13.89\\
         \hline
    \end{tabular}
    \caption{Data Augmentation ASR performance (WER) with reduced source material.}
    \label{tbl:reduced-data}
\end{table}
While this approach is still effective with less source data, we are not able to maintain performance as we reduce the amount of source training data.
We find similar gains to the clean performance when training on 460 hours of human speech, but reduced performance on the noisy test data.  When we reduce the source data to 100 hours, we see a much larger improvement from data augmentation.  Note that in these experiments, we made no model changes to adjust for the amount of training data.  This model failed to converge when trained on 10 hours of speech.  

\section{Lexical Diversity}
\label{sec:ls-lexical-diversity}
One potential advantage of speech synthesis 
is the ability to synthesize unseen utterances thereby expanding lexical as well as acoustic diversity of training material.

\subsection{Topline experiments}
To verify that this approach has merit, we run an optimistic ``topline'' experiment.  Here we assume knowledge of the test utterances, but not the audio.  We then augment the training data with synthesized versions of these utterances using speaker d-vectors sampled from the training utterances or extracted from the test audio.   In the {\bf sampled} condition, there is no test audio used, while in the {\bf original} condition the test audio is used to condition the synthesizer to sound like the test speaker via d-vector extraction.  Results can be found in Table \ref{tbl:ls-topline}.
\begin{table}[htb]
    \centering
    \begin{tabular}{|c||c|c|}
    \hline
    Spkr.~Repr.&test-clean&test-other\\
    \hline
         Sampled&3.2&11.8  \\
         Original& 3.1&11.6 \\
         \hline
    \end{tabular}
    \caption{Topline ASR performance (WER) augmenting training data with test utterances.}
    \label{tbl:ls-topline}
\end{table}
We find, unsurprisingly, that knowledge of the test utterances yields significant ASR gains.  However, it is sufficient to know the lexical content of the utterances, having access to the audio is not necessary.  While this is unreasonable for general purpose ASR, in domain specific applications, there can be a good deal of {\em a priori} information about utterances that users will generate.

\subsection{Language Model Sampling}
To explore lexical diversity without using test data, we generate a MaxEnt language model (LM) \cite{Biadsy2017EffectivelyBT} on the {\sc LibriSpeech} training data.  We use this language model as a generative model to produce new utterances, keeping  those sequences with fewer than 20 words, and perplexity under 500.  We then synthesize these utterances using the {\bf sampled} speaker representation. Filtering out those that are recognized with WER over 20\%, we add them to the {\sc Librispeech} training data.  Table \ref{tbl:ls-lexical-diversity} describes the results of increasing the amount of synthesized utterances. No particular ordering was made to the synthesized utterances, but each set is a subset of each larger set.
\begin{table}[htb]
    \centering
    \begin{tabular}{|c||c|c|}
    \hline
    \# TTS utts& test-clean&test-other\\
    \hline
    0&4.77&13.89\\
    100k&4.80&13.89\\
    600k&4.55&13.64\\
    1.1M&4.70&13.58\\
    \hline
    \end{tabular}
    \caption{Librispeech ASR performance (WER) with lexically diverse data augmentation}
    \label{tbl:ls-lexical-diversity}
\end{table}
We find that increasing the amount of synthesized material up to 600k utterances (approximately the same size as the source training data) results in a 5\% relative reduction of error. This is similar to the finding in Section \ref{ssec:speaker-representation} repeating the training utterances verbatim.  However, we find that there is a limit to the gains from adding additional lexically diverse material.  With 1.1M augmentation utterances, performance is not clearly different from adding 600k synthesized utterances -- we see worse WER on test-clean but better performance on test-other.



\section{Domain Adaptation}
\label{sec:domain-adaptation}

We now explore the use of TTS data augmentation where the TTS domain is distinct from the ASR domain.  The {\sc Isolated-Sentences} material is read, isolated sentences from varied recording conditions, while {\sc LibriSpeech} is made up of audio books, and contains some relatively clean material.


\subsection{Isolated-Sentences data augmentation}
We train a TTS model based on the {\sc Isolated-Sentences} material.  Since the source data is noisy, we find the synthesis quality to be lower.
We then synthesize the {\sc Isolated-Sentences} training data with this TTS model, as well as the {\sc LibriSpeech} TTS model.  Results can be found in Table \ref{tbl:wh-data-aug}.  Since {\sc LibriSpeech} (960hrs) is much larger than {\sc Isolated-Sentences} (76hrs), we also explore ASR training using both corpora.
\begin{table}[htb]
    \centering
    \begin{tabular}{|c||c|c|}
    \hline
    &\multicolumn{2}{c|}{Train Set}\\
    TTS Model&IS&IS+LS\\
    \hline
    None&32.9&30.9\\
    {\sc Isolated-Sentences}&33.0&29.8\\
    {\sc LibriSpeech}&34.2&29.6\\
    \hline
    \end{tabular}
    \caption{ASR results (WER) on {\sc Isolated-Sentences} augmenting either {\sc Isolated-Sentences} or the union of {\sc Isolated-Sentences} and {\sc LibriSpeech} with TTS utterances trained on {\sc LibriSpeech} or {\sc Isolated-Sentences}}
    \label{tbl:wh-data-aug}
\end{table}
We find that training on both data sets results in better performance.  Even though the domains are different, the amount of additional training data from {\sc LibriSpeech} helps performance. However, data augmentation in this condition is able to improve performance further.  The synthesizer used matters somewhat less, with similar performance achieved from both.  However, we see some evidence of domain mismatch when we only train on {\sc Isolated-Sentences} source.  While we do not observe performance differences through augmentation with {\sc Isolated-Sentences} TTS, use of {\sc LibriSpeech} TTS utterances degrades performance.  
 For comparison, unsupervised domain adaptation, where we recognize the {\sc Isolated-Sentences} test utterances with the {\sc LibriSpeech} ASR and fold the hypotheses into the (LS) training data, results in a WER of 35.6\%.

\subsection{Isolated-Sentences lexical diversity}
We then repeat the lexical diversity experiments using a LM trained on the {\sc Isolated-Sentences} corpus.  Since there is less data, we limit the MaxEnt features to 3-grams.  We repeat the topline experiments as well, synthesizing the {\sc Isolated-Sentences} test utterances as well.  Topline results are reported in Table \ref{tbl:wh-topline}.
\begin{table}[htb]
    \centering
    \begin{tabular}{|c||c|c|}
    \hline
    &\multicolumn{2}{c|}{Train Set}\\
    TTS Model&IS&IS+LS\\
    \hline
    {\sc Isolated-Sentences}&24.2&28.9\\
    {\sc LibriSpeech}&24.5&29.3\\
    \hline
    \end{tabular}
    \caption{Topline results (WER) on {\sc Isolated-Sentences} using TTS trained on {\sc LibriSpeech} or {\sc Isolated-Sentences}}
    \label{tbl:wh-topline}
\end{table}
We observe significant gains as expected, but only when training on the {\sc Isolated-Sentences} material alone.  When the training data includes {\sc LibriSpeech}, these gains are not observed.  This may be explained by the ratio of augmentation to training material.  The test material is 5.5\% the size of the {\sc Isolated-Sentences} training data, but only 0.36\% of the combined training data.  

When generating utterances from the LM, we limit these to those with fewer than 5 words, and perplexity under 200.  
Based on findings from Section \ref{sec:ls-lexical-diversity}, we constrain the amount of synthesized utterances to be up to double the amount of source training data, here 400k utterances. Augmentation results are reported in Table \ref{tbl:wh-lexical-diversity}.
%
%
%
%
\begin{table}[htb]
    \centering
    \begin{tabular}{|c||c|c|c|c|}
    \hline
    &\multicolumn{4}{c|}{Train Set}\\

    &\multicolumn{2}{c|}{IS}&\multicolumn{2}{c|}{IS+LS}\\
    \# TTS Utts&IS TTS&LS TTS&IS TTS&LS TTS\\
    \hline
    0&\multicolumn{2}{c|}{32.9}&\multicolumn{2}{c|}{30.94}\\
    100k& 33.53 & 33.02  &29.47&30.67\\
    200k& 31.29 & 33.11  &29.74  &30.21\\
    400k& 30.74 & 32.03 &29.51 &30.34\\
    \hline
    \end{tabular}
    \caption{Isolated-Sentences ASR performance (WER) with lexically diverse data augmentation}
    \label{tbl:wh-lexical-diversity-200ksteps}
    \label{tbl:wh-lexical-diversity}
\end{table}
Here we find that the introduction of synthesized material generated by either synthesizer helps performance.  As in the  {\sc LibriSpeech} experiments, we find similar gains from expanding the acoustic diversity or lexical diversity of the training data.  The use of an in-domain (IS) synthesizer is more helpful than an out-domain (LS) synthesizer.
We find that in-domain TTS-based data augmentation can improve performance from 32.9 to 30.74 WER.  This is roughly equivalent to the gain from including 960 hours of out-of-domain human speech, 32.9 vs. 30.94, with no additional transcribed data.  We can further improve performance to 29.51 using both out-of-domain human speech and in-domain TTS.
\section{Conclusions}

This work has shown methods such that data augmentation from speech synthesis can provide gains to speech recognition.  We improve acoustic diversity by synthesizing training data with different speaker characteristics. We improve lexical diversity by generating new training utterances from an LM trained on the training data.  Both of these approaches result in improved ASR performance.  We then show that these improvements can be observed using a multi-speaker TTS model trained on a distinct domain, here, {\sc Librispeech}, to generate synthetic utterances can improve performance on a target domain.

While we observe improvements in a number of directions in this work, the value of synthesized speech as ASR training data remains dramatically less than that of real speech.

\bibliographystyle{IEEEtran}

\bibliography{tts4asr}

\end{document}